\documentclass[conference,a4paper]{IEEEtran}
\IEEEoverridecommandlockouts
\usepackage{CJKutf8}
\usepackage{cite}
\usepackage{hyperref}
\usepackage{subfigure}
\usepackage{amsmath,amssymb,amsfonts}
\usepackage{algorithmic}
\usepackage{graphicx}
\usepackage{textcomp}
\usepackage{xcolor}
\def\BibTeX{{\rm B\kern-.05em{\sc i\kern-.025em b}\kern-.08em
    T\kern-.1667em\lower.7ex\hbox{E}\kern-.125emX}}
\begin{document}

\title{DeepTracks: Geopositioning Maritime Vehicles in Video Acquired from a Moving Platform}

\author{\IEEEauthorblockN{Jianli Wei, Guanyu Xu, Alper Yilmaz, \textit{Senior Member, IEEE}}
\IEEEauthorblockA{\textit{Photogrammetric Computer Vision Lab., The Ohio State University,} Columbus, OH, USA \\
\textsuperscript{}\big\{wei.909, xu.2186, yilmaz.15\big\}@osu.edu}}

\maketitle

\begin{abstract}
Geopositioning and tracking a moving boat at sea is a very challenging problem, requiring boat detection, matching and estimating its GPS location from imagery with no common features. The problem can be stated as follows: given imagery from a camera mounted on a moving platform with known GPS location as the only valid sensor, we predict the geoposition of a target boat visible in images. Our solution uses recent ML algorithms, the camera-scene geometry and Bayesian filtering. The proposed pipeline first detects and tracks the target boat's location in the image with the strategy of tracking by detection. This image location is then converted to geoposition to the local sea coordinates referenced to the camera GPS location using plane projective geometry. Finally, target boat local coordinates are transformed to global GPS coordinates to estimate the geoposition. To achieve a smooth geotrajectory, we apply unscented Kalman filter (UKF) which implicitly overcomes small detection errors in the early stages of the pipeline. We tested the performance of our approach using GPS ground truth and show the accuracy and speed of the estimated geopositions. Our code is publicly available at \url{https://github.com/JianliWei1995/AI-Track-at-Sea}.
\end{abstract}


\section{Introduction}
\label{intro}
Tracking boats in images is a hard problem especially in oblique viewpoints. Some challenges include partial and full occlusions, small target size especially when the boats are close to horizon line and the area within the image where the boats occur. Adding to these challenges, geopositioning the tracked boat complicates the problem further. This paper proposes a systems approach to geopositioning a moving boat using image sequence acquired from another moving platform.

While there is an influx of papers on object detection and tracking by detection, to the best of our knowledge there is no published literature on geopositioning tracked objects especially for maritime scenarios. By and large, one can apply any object detection and tracking algorithm for image sequences of boats on sea and there are several surveys in the field on their performances. In this paper, we use existing methods, such as \cite{ciaparrone2020deep} and \cite{9142255}, to perform this task. Given the detected set of boats, without the loss of generality, we track the boat that has ground truth GPS information to assess the developed algorithm. 

Estimating the position of the tracked boat requires establishing absolute geometry between the camera and the scene which in this case is the sea. There are two bottlenecks in solving this problem. The first of these is the lack of training data to generate a network solution as in~\cite{kendall2017geometric} where a deep learning framework is used to estimate camera pose. The second problem is the lack of features in the limited sea area that defines the view geometry. In order to mitigate these shortcomings, we model the visible sea region as a plane and estimate the plane projective relation \cite{10.5555/861369} between the camera and the sea. While the earth, hence the sea, has a curvature piece-wise planarity assumption is shown to work well our  experiments. The geopositioning of the boat is followed by application of the UKF to ensure the predicted GPS location produces a smooth sea trajectory. 

The contributions of this paper to geopositioning of boats from image sequences include: 1) being the first paper on geopositioning of sea targets in monocular image sequences; 2) provide a goal oriented neural network modeling the view geometry without traditional learning mechanisms 3) an end to end system composed of detection, tracking and geopositioning modules.

The rest of this paper is organized as follows. Section \ref{sec:related} reviews related work in object detection, tracking and geopositioning. Section \ref{sec:problem} introduces the problem and discusses the details of the proposed approach. Section \ref{sec:experiment} demonstrates the performance and Section \ref{conc} concludes the paper.

\section{Related Work}
\label{sec:related}

Object detection is fundamental problem and active area of research in image processing and computer vision. There have been numerous publications on it over the past few decades, such as two-stage R-CNN series~\cite{girshick2014rich,girshickICCV15fastrcnn,ren2016faster}
and one-stage "you only look once" (YOLO) series~\cite{redmon2016you,redmon2017yolo9000,redmon2018yolov3,bochkovskiy2020yolov4,glenn_jocher_2021_4679653}. These one stage models with pre-trained weights perform well enough for the object detection sub-task in our approach.

Object tracking has also been an active area of research over the past decades. Researchers have proposed numerous methods to track single or multiple objects \cite{he2018twofold, zhu2018online}. In this paper, among alternatives, we choose to apply a tracking by detection (TBD), that associates detected objects across consequent images. As common practice in TBD, we introduce appearance and distance constraints to ensure correct association. The appearance and distance constraints respectively consider the width of the boat and the proximity of the detected tracked object in consecutive frames. 

In this paper, typical to many other practical implementations of object tracking methods, the association in the image space for tracking, and geopositioning the tracked boat in the object space use Bayesian filtering. In particular, the tracking step is accompanied by a Kalman Filter~\cite{jwo2007adaptive, yang2017extended} to model random noise in the object detection step. The geopositioning that generate geo-trajectories uses Unscented Kalman Filter (UKF), which is chosen to model non Gaussian noise that occur due to projective transformation of the detected boat location. The UKF uses the speed and direction as object state and models their change over time.

\section{Methodology}
\label{sec:problem}

Geopositioning an object tracked in a monocular image sequence introduces a number of challenges that include precise tracking of the boat in the image plane (pixel errors translate to large geoposition errors); defining a geometric relation between image and GPS coordinate systems; an end-to-end system that resolves this problem; reduction of geoposition errors due to large oblique view point.

\begin{figure}[htbp!]
\centerline{\includegraphics[width=0.9\columnwidth]{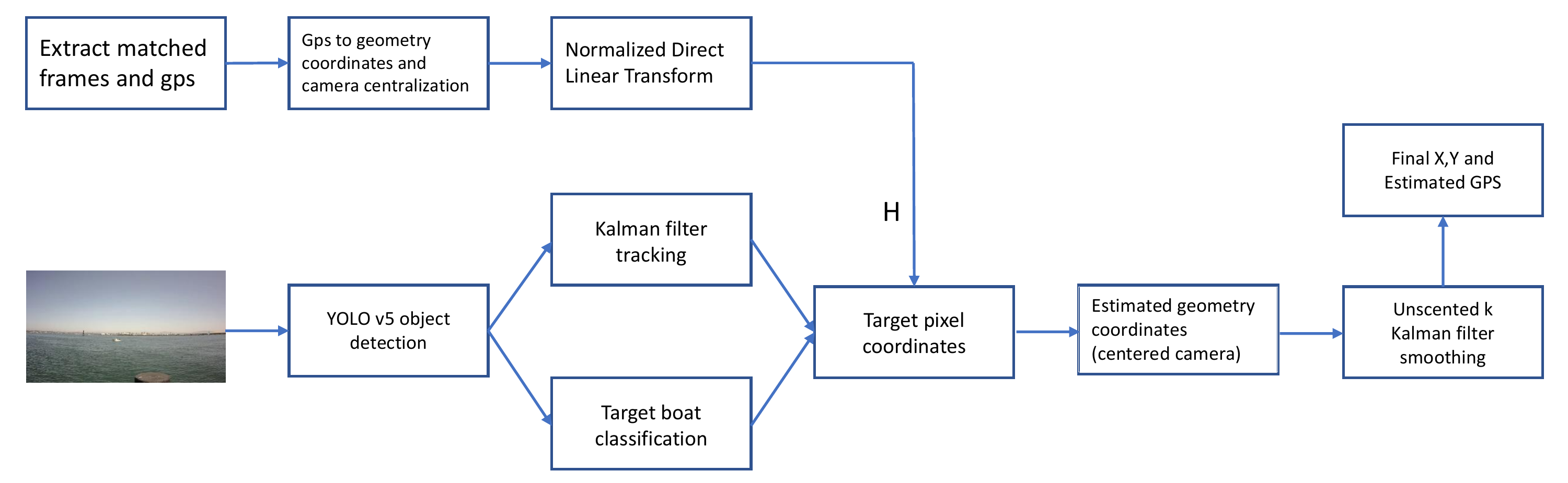}}
\caption{Flow diagram showing the relation between the geometric, detection, classification, tracking and geopositioning modules.}
\label{Flowchart}
\end{figure}

To address these challenges, the proposed end-to-end system uses a cascade of neural networks that is composed of several modules: the coordinate transformation module, detection module, classification module, tracking module and post processing module illustrated in Figure \ref{Flowchart}. The first module is a linear Multilayer Perceptron (MLP) that performs the geometric transformation between the image and the world reference frames. This is followed by an boat detection module which also contributes to the generation the MLP. The method proceeds by verification of the detected boat using a CNN classifier, and Kalman filtering in both the image and world coordinates to generate tracks and the geoposition of the target boat.

\subsection{Coordinate transformation}
\label{CooTrans}

\begin{figure}[htbp!]
\centerline{\includegraphics[width=\columnwidth]{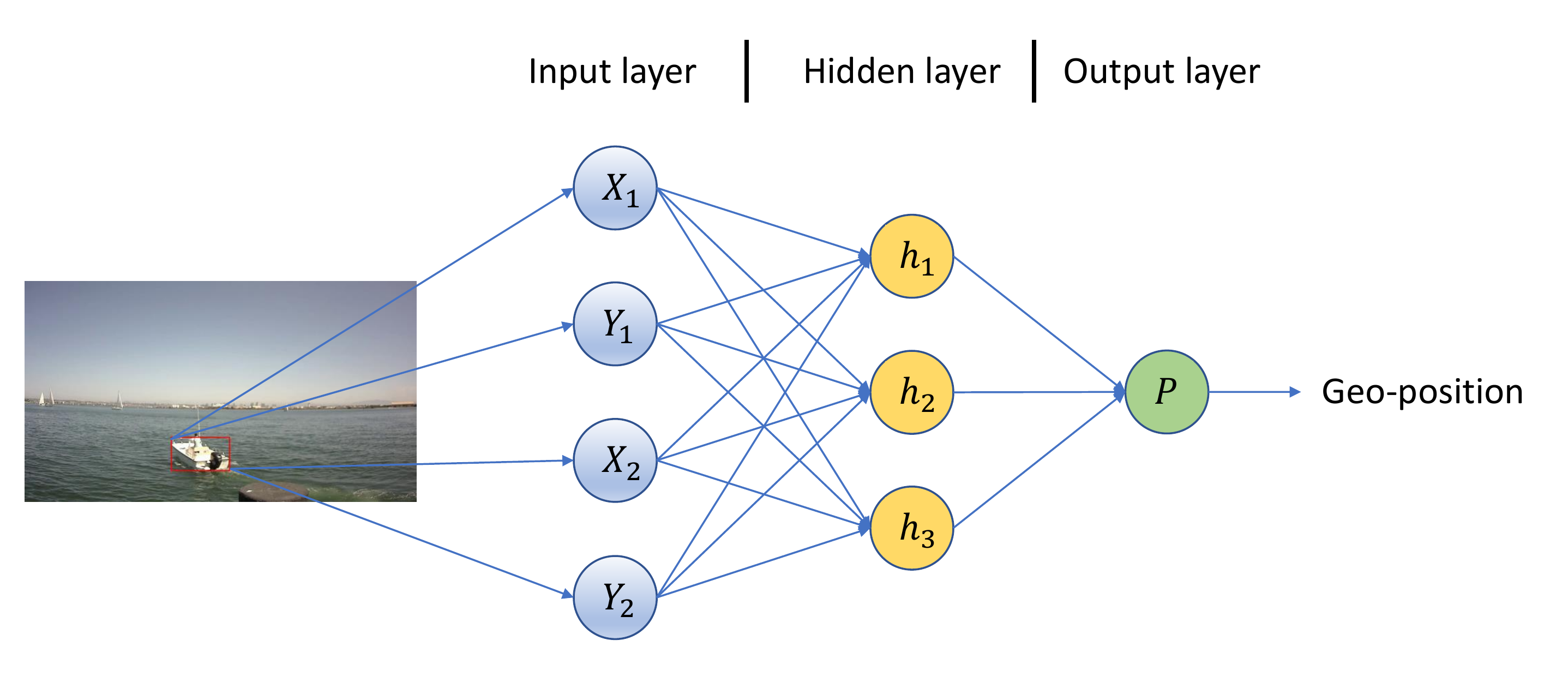}}
\caption{Linear MLP modeling the geometric transformation between the camera and world reference frames.}
\label{MLPs}
\end{figure}

Obtaining position in the world frame from a monocular camera is not a trivial task. First and foremost, the developed approach should include a methodology to transfer the boat location in the video to GPS location in world reference frame. In order to achieve this goal, we introduce a linear MLP based geometry model that transfers pixel location to absolute sea coordinates which is modeled as piece-wise plane. In this model, the MLP parameters are estimated using the training set provided with the dataset which contains the image position $(x,y)$ of a boat and its corresponding GPS position $(lat,lon)$. Using this quadruplet $<(x,y),(lat,lon)>$, we set the north and east directions of the camera as \textit{x} and \textit{y} axes of the quasi sea-plane that overlap with the GPS coordinates in western hemisphere. Since, the variation of latitude and longitude as the boat sails remain infinitesimally small over time, we convert the latitude and longitude differences to meters on the sea-plane. The boats in the image are detected using a trained version of the YOLO object detector from which the bottom center of the boat's bounding box that lies on the sea surface is used as the boat location. This and its corresponding GPS location is then used to compute the plane-projective transformation that estimates the homography matrix \textit{H} between the two coordinates:
\begin{equation}
    T_{world}*X_{world} = H*T_{local}*X_{local}
\end{equation}
\noindent where $T_{world}$ and $T_{local}$ are normalization matrices that serve as preconditioning transformations to decrease condition number of H, $X_{world}$ and $X_{local}$ pairs are respectively homogeneous world geometry coordinates and homogeneous local pixel coordinates. The resulting H matrix along with the lens distortion parameters of the camera are converted into a linear MLP as shown in Figure \ref{MLPs}.

\subsection{Object detection and tracking}
\label{ODT}

The detection of the boats in the video sequences uses  YOLO~\cite{glenn_jocher_2021_4679653}, which is arguably one of the most advanced object detection algorithms that can detect multiple objects in real time. In our implementation, we utilized YOLO to detect only ‘boats’. Given the detected boats in consecutive frames, tracking is accomplished by a two stage Kalman filtering operation. The first stage in Kalman filtering operates on the consecutive images that ensures consistent boat trajectory and reduce object detection errors that are assumed Gaussian. It uses the output from the object detector as the observations used to correct the boat state ($X_{local}$,$\hat{X}_{local}$ ). In particular, we use the boat position \textit{(x,y)} and the boat width (\textit{w}) as observations in the image frame. Besides, an additional VGG-16~\cite{simonyan2014very} based boat classifier trained using the training data improves methods robustness by assisting the Kalman filter and ensures proximal and similar boat is associated to the boat in the previous frame. We compute a detection score by convexly combining the classification network probability $P_c$ and Kalman filter distance distribution $P_K$:
\begin{align}
    & P_K = e^{-\frac{d^2}{\sigma^2}} \\
    & P = \alpha*P_C+(1-\alpha)*P_K
\label{disteq}
\end{align}
where \textit{d} is predicted target boat distance between two continuous frames, $\sigma$ is the distance standard derivation, $\alpha$ is the combination parameter and \textit{P} is the combined probability factor for the tracked target boat.

\begin{figure}[htbp!]
\centerline{\includegraphics[width=1.0\columnwidth]{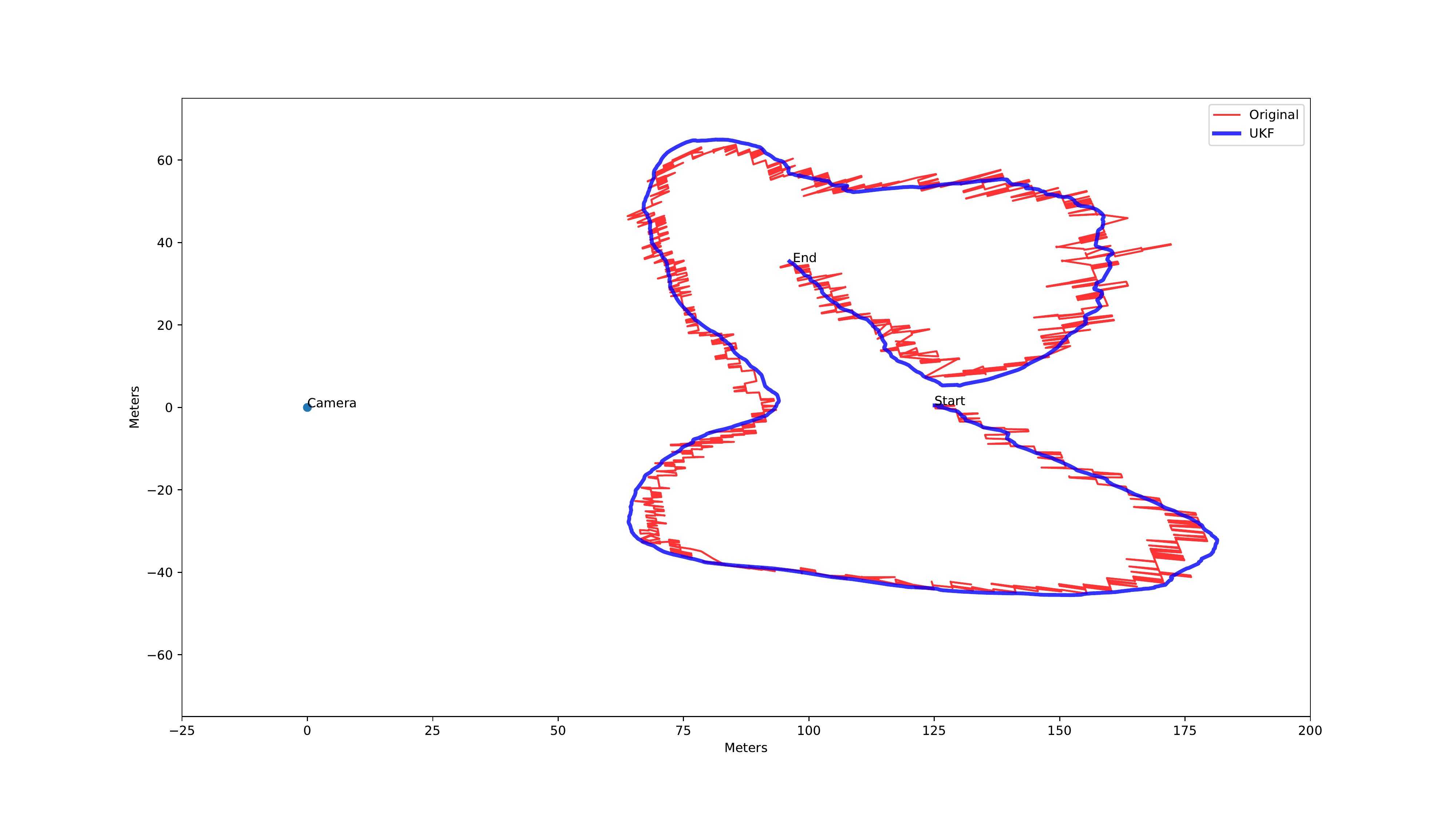}}
\caption{Predicted raw geo-trajectory and processed geo-trajectory using UKF in the world frame.}
\label{Traj_UKF}
\end{figure}

The second stage Kalman filter operates in the world frame to generate consistent latitude longitude information. For both stages, the boat’s states are composed of positions $X_{local}$ and $X_{world}$ as well as the speeds $\hat{X}_{local}$ and $\hat{X}_{world}$ in respective coordinate systems. This Bayesian filtering is necessary to reduce problems related to the weak geometric relation between the image and world frames. The weak geometry is due to small support area where the  quasi-collinear boat locations in the training dataset occur and that only provides observations close to the horizon. Hence, any small error, such as 1 pixel, in the image contributes to meters of error in the world frame. This error can be reduced to a certain degree by the second stage Kalman filtering which provides smooth boat trajectories. In Figure \ref{Traj_UKF}, we show the predicted boat trajectory in the world frame before/after the application of the second stage UKF.

\section{Experiments}
\label{sec:experiment}

In the training process, we trained a boat classifier for classifying target boat by appearance only and use the GPS positions in the training set to estimate the geometric relation modeled as a linear MLP. The experiments of the proposed system uses the dataset provided by the Office of Naval Research.

\subsection{Dataset}
\label{subsec: datasets}
The dataset used for evaluation of the proposed approach includes monocular image sequence that contains GPS of a target boat and geoposition of the camera. The sequence contains multiple boats in motion within the field of camera view, including the target boat which the proposed approach seeks to distinguish, track and estimate a geoposition. Target boat GPS data provided every 1 to 5 seconds is used as ground truth. The camera used to acquire the sequences is on a moving platform and has an associated GPS information. Dataset is available at \url{https://drive.google.com/drive/folders/1Eq7afvav49OmWo5iNSKL7fEJk-TGNp0V}.

\subsection{Implementation details}
\label{subsec: implementation}

In our experiment, for target boat tracking, we introduce a probability threshold, such that we considered it as our target boat and the algorithm continues tracking if combined probability \textit{P} exceeded the threshold. We preset the threshold $P_{thr}$=0.51, combination coefficient $\alpha=0.5$ and distance standard derivation $\sigma=10$. No matter classifier probability or Kalman filter approaches to 1, the target boat would be detected and tracked. When the boat is recognized by our classifier and the Kalman filter is over $P_{thr}$, we can ensure the reappeared boat is the target boat. 

\subsection{Results}
\label{subsec: experiment results}

\begin{figure}[htbp]
\centering    
\subfigure[]{
\label{fig:4a}
\includegraphics[width=0.95\columnwidth]{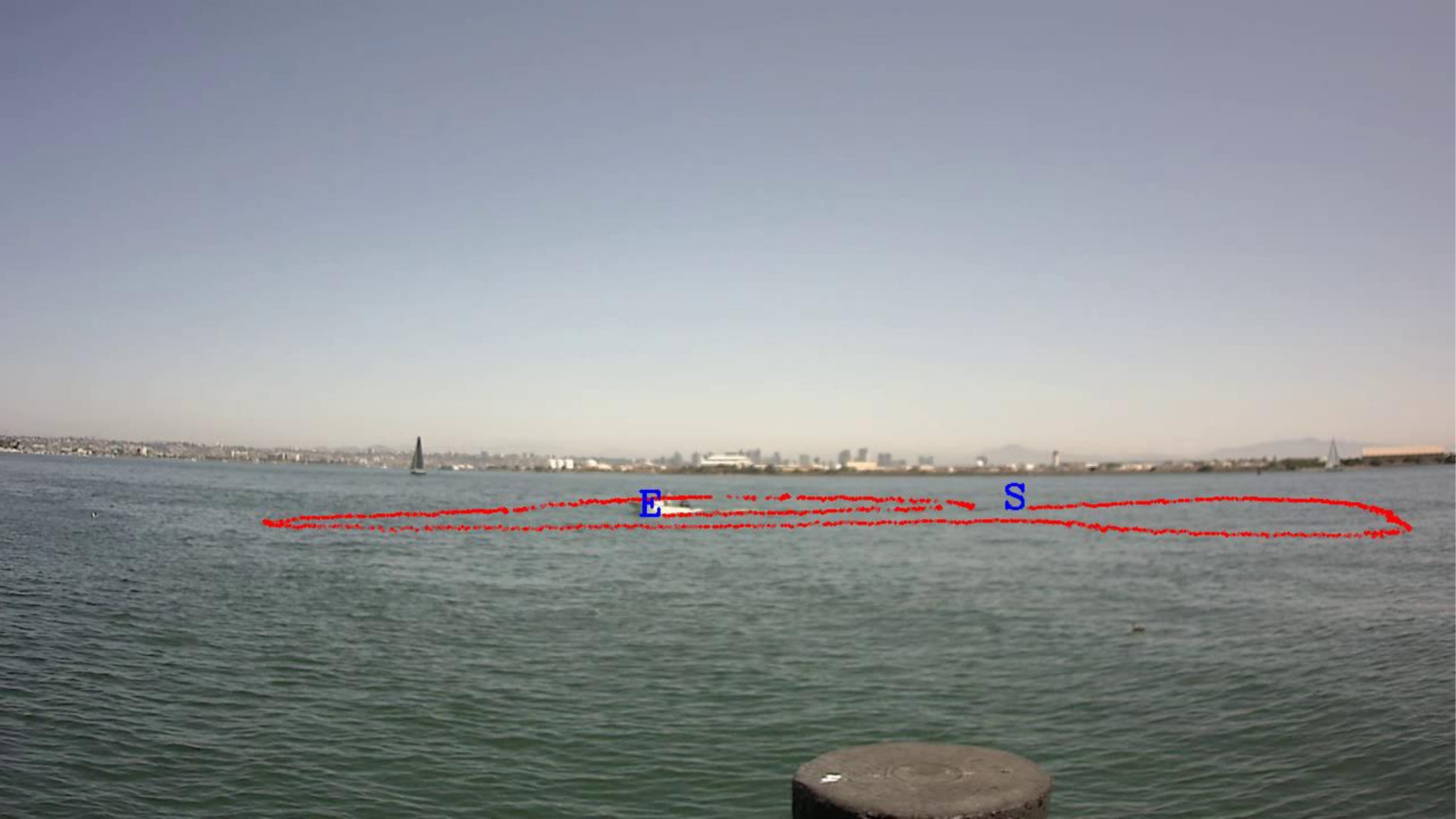}}
\hspace*{-4mm}
\subfigure[]{
\label{fig:4b}
\includegraphics[width=1.05\columnwidth]{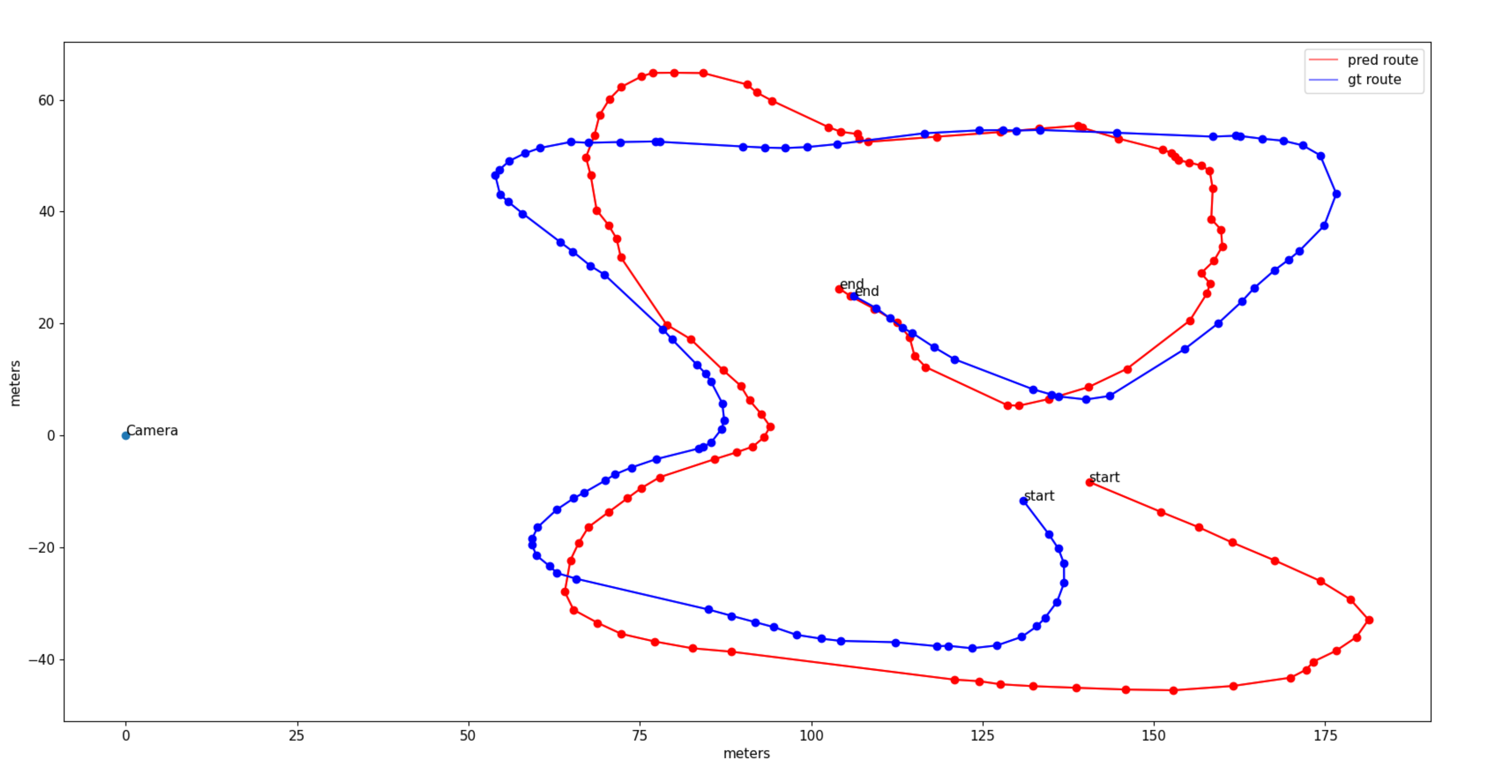}
}
\caption{(a) Boat trajectory on the video frame, ‘S’ and ‘E’ indicates where the boat start and end its trajectory. (b) Predicted boat trajectory in the world frame plotted against its ground truth with position denoted as blue dot.}     
\label{result1}     
\end{figure}

In Figure \ref{result1}, we show the detected and tracked boat location in the image plane \ref{fig:4a} and the predicted boat location in the world frame plotted against the ground truth GPS position provided in the dataset \ref{fig:4b}.

\section{Conclusions and Future Work}
\label{conc}

Our AI and monocular camera based approach aims to estimate target boat gps location at sea from the video. We noticed that the Kalman filter was not capable to cope with some extreme conditions and tracking module shifts the target to other neighboring boats. In the future, we will tackle this problem and proposed a multi-boats track at sea that could predict multiple boats GPS locations simultaneously.

\bibliographystyle{ieeetr}
\bibliography{Main}

\end{document}